\documentclass[sigconf]{acmart}
\AtBeginDocument{%
  }

\usepackage{natbib}
\usepackage{algorithm}
\usepackage{algorithmic}
\usepackage{booktabs}
\usepackage{balance}

\copyrightyear{2025}
\acmYear{2025}
\setcopyright{cc}
\setcctype{by-nc-sa}
\acmConference[SIGIR '25]{Proceedings of the 48th International ACM SIGIR Conference on Research and Development in Information Retrieval}{July 13--18, 2025}{Padua, Italy}
\acmBooktitle{Proceedings of the 48th International ACM SIGIR Conference on Research and Development in Information Retrieval (SIGIR '25), July 13--18, 2025, Padua, Italy}\acmDOI{10.1145/3726302.3731959}
\acmISBN{979-8-4007-1592-1/2025/07}

\setcopyright{none}
\renewcommand\footnotetextcopyrightpermission[1]{}
\begin{document}

\title{Insight Agents: An LLM-Based Multi-Agent System for Data Insights}

\author{Jincheng Bai}
\affiliation{%
  \institution{Amazon}
  \city{Seattle}
  \country{USA}}
\email{baijin@amazon.com}

\author{Zhenyu Zhang}
\affiliation{%
  \institution{Amazon}
  \city{Seattle}
  \country{USA}}
\email{zhenyuzh@amazon.com}

\author{Jennifer Zhang}
\affiliation{%
  \institution{Amazon}
  \city{Seattle}
  \country{USA}}
\email{yezhan@amazon.com}

\author{Zhihuai Zhu}
\affiliation{%
  \institution{Amazon}
  \city{Seattle}
  \country{USA}}
\email{jasonzhu@amazon.com}

\thanks{This is the author's version. Published in \textit{Proceedings of the 48th International ACM SIGIR Conference on Research and Development in Information Retrieval (SIGIR '25)}, \url{https://doi.org/10.1145/3726302.3731959}}

\renewcommand{\shortauthors}{Jincheng Bai, Zhenyu Zhang, Jennifer Zhang, and Jason Zhu}

% REMOVE THIS: bibentry
% This is only needed to show inline citations in the guidelines document. You should not need it and can safely delete it.
% \usepackage{bibentry}
% END REMOVE bibentry

\begin{abstract}

Today, E-commerce sellers face several key challenges, including difficulties in discovering and effectively utilizing available programs and tools, and struggling to understand and utilize rich data from various tools. We therefore aim to develop Insight Agents (IA), a conversational multi-agent Data Insight system, to provide E-commerce sellers with personalized data and business insights through automated information retrieval. Our hypothesis is that IA will serve as a force multiplier for sellers, thereby driving incremental seller adoption by reducing the effort required and increase speed at which sellers make good business decisions. In this paper, we introduce this novel LLM-backed end-to-end agentic system built on a plan-and-execute paradigm and designed for comprehensive coverage, high accuracy, and low latency. It features a hierarchical multi-agent structure, consisting of manager agent and two worker agents: data presentation and insight generation, for efficient information retrieval and problem-solving. We design a simple yet effective ML solution for manager agent that combines Out-of-Domain (OOD) detection using a lightweight encoder-decoder model and agent routing through a BERT-based classifier, optimizing both accuracy and latency. Within the two worker agents, a strategic planning is designed for API-based data model that breaks down queries into granular components to generate more accurate responses, and domain knowledge is dynamically injected to to enhance the insight generator. IA has been launched for Amazon sellers in US, which has achieved high accuracy of 90\% based on human evaluation, with latency of P90 below 15s.
% As one of use cases, the IA has been announced as a highlight to all Amazon sellers in September 2024. 

\end{abstract}



\maketitle

\section{Introduction}
%Amazon today can offer customers a wide range of products at competitive prices thanks to its vast network of over 2 million active Sellers across 200+ countries. However, Sellers on Amazon face several key challenges: (1) Struggle to extract insights from data and prioritize issues across multiple fragmented tools (10+ dashboard/reports, 40+ growth level actions, etc.). (2) Lack of personalized 1:1 support and guidance for growing their business.

%Data shows that Sellers with human guidance (e.g., Growth Consultants) to help them access data and make decisions drive 2X higher adoption rates compared to self-serve workflows. This gives confidence to roll out a more scalable solution powered by Generative AI (GenAI). This solution aims to revolutionize seller experiences by offering interactive, personalized built-in capabilities under data insights pillar. It not only fosters increased seller adoption of existing tools (projected 8\% increase), unlocking growth potential (projected incremental \$3B annualized opportunity), but also lays the foundation for answering Seller-related queries on Amazon's centralized help/search assistant for Accelerate 2024.

% It takes a lot to run a small business, from designing and manufacturing products to delight customers, to hiring and advertising.
Running a small business involves numerous challenges, from product design and manufacturing to marketing and advertising.
As an E-commerce company, 
% we support sellers with powerful, cost-effective tools that incorporate the latest advancements in optimization and machine learning to help sellers identify unmet customer needs, forecast demand for their products, manage inventory levels, determine optimal pricing, and much more. 
we support sellers with powerful tools that leverage advanced machine learning to optimize pricing, forecast demand, manage inventory, identify market opportunities and many more.
Although we provide sellers with rich tools to manage the facets of their business, it can be a challenge for some sellers to find the right tool, and leverage insights effectively. To remove these barriers, we develop Insight Agents (IA), which is an LLM-based \cite{Wei2022Emergent} multi-agent conversational assistant, providing personalized data and business insight through automated information retrieval. It aims to reduce sellers’ cognitive load and unlock their potential to grow business.

Specifically, IA enables seller to talk to their data 
% by answering “what” questions, providing responses in narrative or simple visual and tabular format addressing sellers’ challenges of discoverability and complexity; in addition, it allows sellers to gain deeper insights to their business through benchmarks, identifying strengths and weakness and uncovering trends and opportunities for growth and optimization. Particularly, we prioritize two types of seller requests: 
through two main types of requests: (1) Descriptive Analytics that presents data according to the specified query. Examples include “what were my sales and traffic for the top 10 products last month”, and “how does my monthly sales change year over year”. (2) Diagnostic Analysis encompassing summarization, benchmarking, and other analytical techniques. Examples include “how does my business perform”, “how is my business doing with respect to my benchmarks”.

Building a reliable and helpful IA system is challenging, as it requires simultaneous consideration of coverage, accuracy and latency. 
% However, it poses great difficulties to build a reliable and helpful IA with simultaneous consideration of coverage, accuracy and latency. 
To circumvent these challenges, we propose an end-to-end agentic workflow built on a plan-and-execute paradigm and designed for comprehensive coverage, high accuracy, and low latency. It employs a hierarchical manager-worker multi-agent structure to optimize data retrieval and question answering, which follows the general multi-agent concept \cite{wang2024agents, Durante2024agent} utilizing different resolution paths tailored to the query type. The \textbf{manager agent} mainly consists of Out-of-Domain (OOD) detection and branch routing. 
% We propose two-layers Out-of-Domain (OOD) detection to respond quickly to out-of-scope questions without sacrificing coverage. 
We build specialized lightweight models for OOD (an encoder-decoder based detector) and branch routing to resolution path to optimize latency (a BERT-based classifier). We then divide the solution space into two worker agent-based resolution paths, \textbf{data presenter agent} and \textbf{insight generator agent}, to facilitate problem-solving. 
Essentially, IA is built upon the Retrieval-Augmented Generation (RAG) \cite{Gao2024RAG} framework, leveraging seller data in a tabular format through a robust and strategic API-based data model. The system decomposes queries into appropriate grains based on company's internal data API availability in a divide-and-conquer manner, with automated tabular data retrieval and aggregation facilitated through task decomposition, planning, and API/function selection. This process, powered by LLM, is analogous to LLM-based task planning \cite{Paranjape2023art, Li2024stride} and tool selection \cite{song2023restgpt, Qin2024toolllm}. 
% similar to text-to-SQL solutions \cite{Gao2023texttosql, Li2023RESDSQL, Liu2023comprehensive, Pourreza2023DINSQL}.
% To ensure accurate response, we propose a robust and strategic API-based data model which decomposes the query into appropriate grains based on data API availability and in a divide-and-conquer manner. 
In addition, we inject domain knowledge dynamically based on query to make insight generator domain-aware. %Our primary data sources are readily accessible through Seller Central via API, and details can be found in the Appendix. 
Overall, we implement the manager-worker multi-agent agentic workflow with orchestration architecture to streamline the entire process, encompassing query understanding, information retrieval and answer generation. IA has been launched for Amazon sellers in US, which has achieved accuracy of 90\% based on human evaluation, with latency of P90 below 15s.

\section{Methodology} \label{sec:method}
% In this section, the end-to-end RAG framework with orchestration for the growth LLM is introduced. The data API with function calling solution for tabular data retrieval is also elaborated.
% In this section, we at first introduce overall hierarchical manager-worker multi-agent architecture, design for its two resolution path (data presenter and insight generator) and design considerations. Then we provide more details about each component in IA which we modularize and are shared among the overall IA architecture. 

\subsection{Architecture Design}

\begin{figure}
  \centering
  \includegraphics[width=8cm]{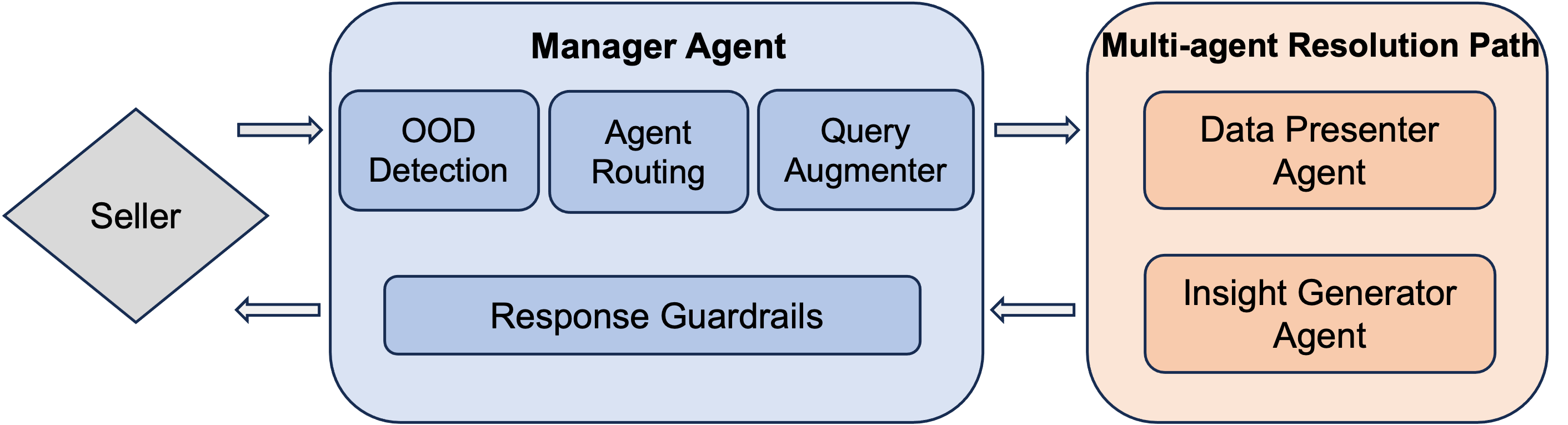}
  \caption{Overall IA Architecture. It illustrates the overall hierarchical structure, consisting of a manager agent overseeing two subordinate worker agents: data presenter agent and insight generator agent.}
  \label{fig:branching}
\end{figure}

\textbf{Overall IA Architecture.}
The high-level hierarchical manager-worker multi-agent architecture of IA is illustrated in Figure \ref{fig:branching}. Upon receiving a seller's query, the manager agent first checks its eligibility against the scope of data insight via Out-of-Domain (OOD) component. It also includes the agent routing component to select the appropriate solver, and the query augmenter to de-ambiguate the query based on its best knowledge.
We design two distinct agent-based resolution paths, i.e., data presenter and insight generator agent which are separable in terms of the solution they provide. Before returning response to sellers, IA applies guardrail to prevent response that contains issues such as  PII leakage, toxic message, etc, from exposing to the seller as the post-processing step.

% \begin{figure*}
%   \centering
%   \includegraphics[width=15cm]{low_level_6.png}
%   \caption{Architecture design for data presenter and insight generator. }
%   \label{fig:components}
% \end{figure*}

\begin{figure}
  \centering
  \includegraphics[width=8cm]{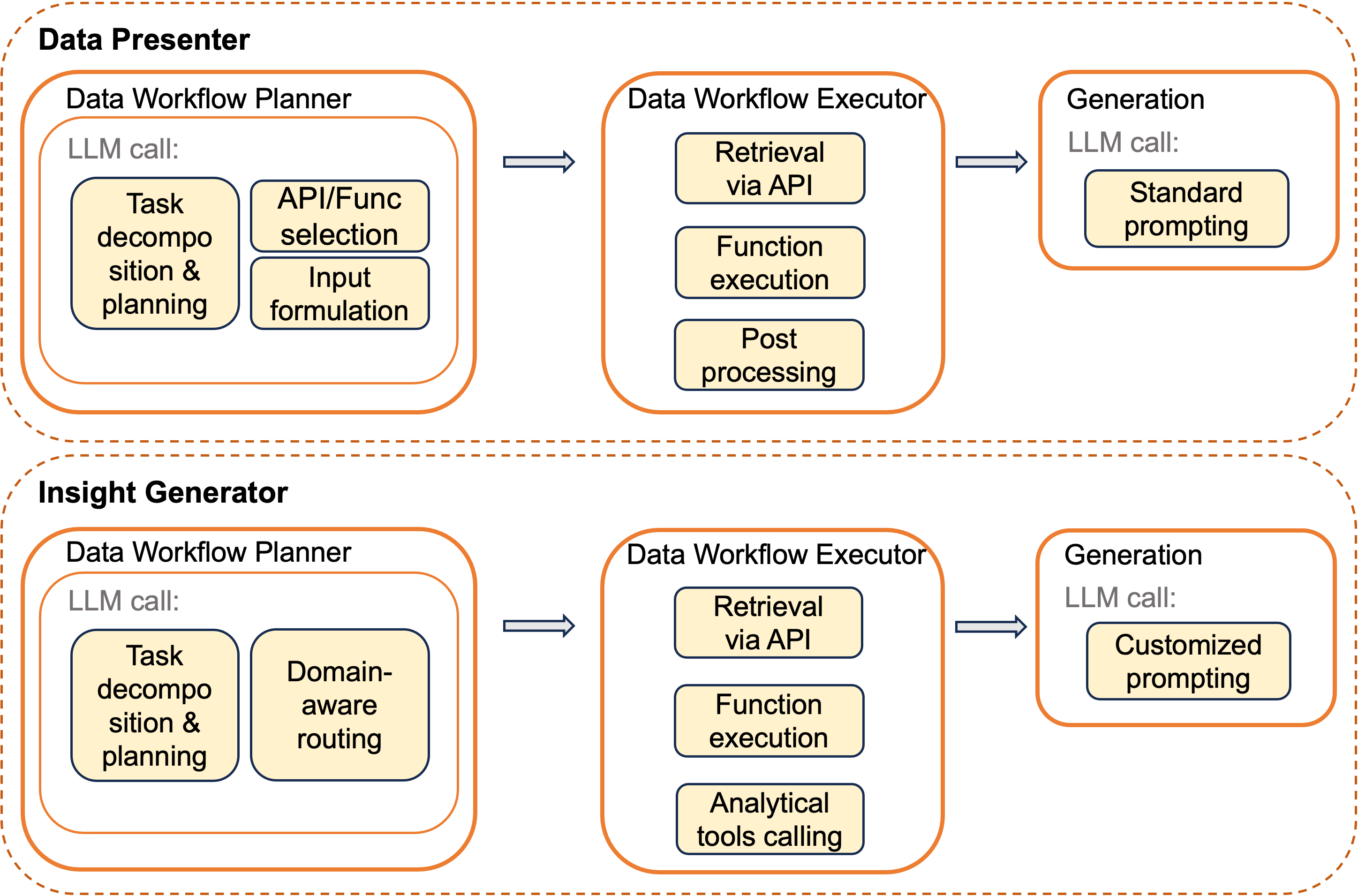}
  \caption{Architecture design for data presenter and insight generator. }
  \label{fig:components}
\end{figure}

\textbf{Design for Data Presenter and Insight Generator.}
Figure \ref{fig:components} illustrates the low-level architecture design of the IA, specifically elucidating the details of the two agents in the resolution paths, which will be explained in section \ref{sec:worker}.
% For both resolution paths, we propose a robust and strategic Data Workflow Planner to retrieve and aggregate data. Specifically, it leverages API to retrieve data which essentially adds structure or constraint due to specific API invocation format, thus trading off flexibility for accuracy as compared to text-to-SQL solution. In addition, Data Workflow Planner decomposes query into steps (grains) based on available data APIs to retrieve data and applies divide-and-conquer (DC) based task decomposition for multiple API calls. The different objectives of the two agents necessitate the employment of distinct data planner for each in terms of data specificity. In particular, data presenter requires customized transformations and aggregation in the data planner to adequately address the diverse range of presentation-related inquiries. In contrast, the domain-aware insight generator necessitates mapping retrieved data to domain knowledge which subsequently invokes a set of predetermined analytical tools and inject contextual knowledge to prompt LLM. However, data retrieval and aggregation module are shareable between two agents, followed by the customizable analytical tools.

\textbf{Implementation Considerations.}
One bottleneck for LLM-based agents is the latency. To overcome this, there are two considerations to address it: model-based components and parallelization. Specifically, we develop specialized model to avoid unnecessary LLM calls while ensuring high performance and only apply LLM for components that involve data fusion and generation. Regarding parallelization, OOD detection and agent routing are performed simultaneously, with both the data presenter and insight generator branches being initiated concurrently with early termination. The parallelization approach trades off increased infrastructure and computation costs for latency reduction.
% We also apply early termination to reduce the additional cost overhead.

% \begin{figure*}
%   \centering
%   \includegraphics[width=15cm]{data_presenter_gen_3.png}
%   \caption{Data Presenter Examples. The numbers provided are for illustrative purposes only and do not represent actual data.}
%   \label{fig:dp_gen}
% \end{figure*}
% \begin{figure}
%   \centering
%   \includegraphics[width=8cm]{data_presenter_gen_3.png}
%   \caption{Data Presenter Examples. The numbers provided are for illustrative purposes only and do not represent actual data.}
%   \label{fig:dp_gen}
% \end{figure}

\subsection{Manager Agent}
% Manager agent orchestrates the end-to-end query answer process via qualifying the query and selecting the appropriate worker agent. Upon receiving the response from the worker agent, the manager agent applies necessary guardrails to ensure the output adheres to established guidelines and constraints.

\textbf{Out-of-Domain (OOD) Detection.}
The out-of-domain (OOD) detection within manager agent serves as the first gating after receiving the query. The component understands seller intent to determine whether it is out of scope for the data insight agent to answer. The in-scope is defined as at least part of the question can be answered based on the available data. 
% We aim to design a high-precision classifier which favors precision over recall of the problem space for data insight agents. The rationale is that this OOD serves as the first-layer guardrail to generate quick response without running through full IA processes for out-of-scope question, which thus optimizing latency. For false negative example, we can still detect in data presenter or insight agent agent for various out of scope reason e.g., due to data availability. 
We create a high-precision classifier that favors precision over recall when filtering data insight requests, to act as an initial screening layer without running the full analysis pipeline. While some invalid requests may not be filtered out (false negatives), they can still be caught later in the data workflow planing stage.
Specifically, we implemented Auto-encoder (AE) based OOD detection, similar to \cite{Torabi2023AE} to depict the boundary of in-scope problem space. In particular, the Auto-encoder with one hidden layer is trained with embedding from sentence transformer \cite{reimers-2019-sentence-bert}. Denote the input embedding vector as $X$, and the encoder transforms it to a hidden representation $H$ as
% \begin{equation}
%     H= \sigma(W_{1}X + b_{1})
% \end{equation}
$H= \sigma(W_{1}X + b_{1})$ and followed by the decoder,
% \begin{equation}
%     \hat{X}= \sigma(W_{2}H + b_{2})
% \end{equation}
$\hat{X}= \sigma(W_{2}H + b_{2})$
with weights $W_1, b_1, W_2, b_2$. The AE is trained to minimize the reconstruction error $r = ||X-\hat{X}||$. We train the AE on the the set of in-domain questions denoted as $\mathcal{X}_{id}$, and the threshold for OOD is determined by 
\begin{equation}
    \mu_{id} + \lambda * \sigma_{id}
\end{equation}
where $\mu_{id}$ and $\sigma_{id}$ denotes the mean and standard deviation of the reconstruction loss $r$ on $\mathcal{X}_{id}$. The hyper-parameter $\lambda$ controls the precision v.s. recall in the OOD detection. 

\textbf{Agent Router.} 
% \textbf{First-Level Branching for Resolution Path Selection:} 
The agent router solves a classification problem by categorizing input queries between data presenter and insight generator. To strike a balance between latency and accuracy, a lightweight transformer-based model could be employed, ensuring responsive performance while maintaining robust classification capabilities. In particular, we fine-tuned a lightweight BERT \cite{Devlin2016bert} based model (33M parameters) on super-sampled data, encompassing presenter and insight generator questions with variations. 
% The branch routing module is an BERT-based classifier to select between data presenter and insight generator. The BERT model is fine-tuned on over-sampled data, encompassing presenter and insight generator questions with variations.

% \textbf{Second-Level Branching for Insight Generator:} In addition, within the insight generator module, there is another layer of branch routing process according to the relevant data sources based on the user query, achieved via few-shot learning based LLM classifier. In particular, the query could be routed to insights derived from domain-specific data. The routing process determines the resolution paths, which consist of the following: (1) the associated analytical analysis that will be incorporated into the context, where analytical analysis may include data aggregation, time series based seasonal \& trend analysis, as well as benchmark analysis among other techniques. (2) the prompt template and few-shot examples for the final insight generation. 

% It is worth noting that as we expand to incorporate more data sources, the static prompt template and few-shot example assignments powered by branching can be enhanced to dynamic assignments by leveraging the Retrieval-Augmented Generation (RAG) solution. In this approach, the prompt template and exemplars are dynamically retrieved from a knowledge base according to the query.

\textbf{Query Augmenter.}
It clarifies, rewrites, and expands queries to reduce ambiguity. In particular, the ambiguity regarding data insight related questions often centers around the time range specified. For example, question "\textit{What were my sales for the last week?}"
% % or 
% "\textit{What were my traffic for the last month?}"
lacks clarity like what the current date is, and the LLMs might also have difficulty accurately interpreting time ranges like "last week". 
% or "last month".
 In this step, contextual information like today's date, the start and end dates of the current week
% , and the current quarter 
are dynamically injected with specific instructions like "week" referring to the calendar week. This step augments the prompt for LLM call in the subsequent step.

% doesn't involve any LLM call, and augmented prompt with contextual information will be utilized in the subsequent step.

% \textbf{Response Guardrails.}
% The guardrail is the post-process step that prevents response that contains issues such as  PII leakage, toxic message, etc, from exposing to the seller. The guardrail is implemented by few-shot learning with LLM.

\subsection{Data Presenter and Insight Generator Agents}\label{sec:worker}
% In this section, we describe the shareable components for Data Presenter and Insight Generator Agents and elaborate on their nuanced difference in each key components.
We examine the shared components of Data Presenter and Insight Generator Agents, highlighting their key differences.

\subsubsection{Data Workflow Planner: a Robust Data Model}

\begin{figure*}
  \centering
  \includegraphics[width=17.7cm]{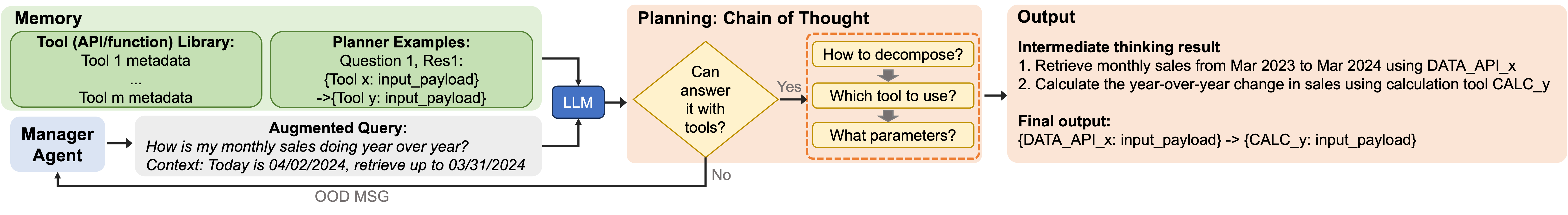}
  \caption{Illustration for data workflow planner. Use data presenter as an example.}
  \label{fig:demo_data_presenter}
\end{figure*}

Our modeling framework relies on retrieval-augmented generation (RAG) technique \cite{Gao2024RAG}, as the insight related responses are grounded on seller data. As specific to IA, the external resources are stored in tabular form, which necessitates the development of tabular retrieval method different from retrieving from unstructured text data. To ensure accuracy of data retrieval and aggregation, we propose a robust data model to retrieve data that leverages company's internal data APIs, decomposes the query into solvable steps based on API availabilities, following a Divide and 
Conquer manner. 
% Concretely, it utilizes API to retrieve data, which inherently imposes structure or constraints.
It specifically uses APIs to retrieve data, which naturally imposes structure and constraints but yields higher accuracy. This represents a trade-off between flexibility and precision when compared to text-to-SQL solutions \cite{Gao2023texttosql, Li2023RESDSQL, Liu2023comprehensive, Pourreza2023DINSQL}, which additionally require the effort of creating and maintaining a relational database.
% which essentially adds structure or constraint 
% due to specific API invocation format, thus trading off flexibility for accuracy. In regular task decomposition, the subtasks are interweaving and conducted sequentially. To ensure the LLMs to be less prone to the intermediate errors, for data retrieval, we propose to decompose the subtasks which are decoupled from each other and then merge the result from subtasks. DC is feasible to data insight agent task since the data retrieval can be formulated as the homogeneous sub-tasks with smaller problem sizes. 
Besides data APIs, leveraging external calculation tools for data transformation is also crucial as calculation errors remain a common challenge for LLMs \cite{Li2024Math, Ahn2024Math}.
The whole process is similar to tool learning and usage by LLM \cite{qu2024tool}, with few-shot examples and tool metadata stored in memory. Workflow planner encompasses task decomposition and planning, along with the selection of appropriate APIs/functions and payload generation with slot filling.
% , with the overarching goal of effectively answering queries by leveraging the existing APIs/functions.
An end-to-end illustration for data workflow planner can be found in Figure \ref{fig:demo_data_presenter}.
% The prompt templates for the Data Workflow Planner component can be found in the Appendix \ref{fig:prompt_api}.

% The API solution’s drawback lies in its inflexibility to accommodate varied seller questions, as the output from the data fetching API adheres to a fixed format. We address this limitation by incorporating aggregation functions and tool invocations via LLM, coupled with query-based task decomposition and planing.

% Tool Learning and usage by LLM \citet{qu2024tool}. Task decomposition and planning, data retrieval via API selection and API payload generation (code generation), calculation tool selection and payload generation. Can be extended SQL code generation. It is achieved in one LLM call.

\textbf{Data-based Out-of-Scope Detection.} 
% Through this step, the LLM leverages the provided dataset to facilitate out-of-scope detection for incoming queries, ensuring that any requests falling outside the intended scope are accurately identified based on the available data. In the prompt, the metadata of the available data sources are provided for detection, and an "out" option like "\textit{The question is out of scope!}" was provided to prevent hallucinations.
The LLM performs secondary out-of-scope detection by comparing incoming queries against provided dataset metadata. An explicit "out" option is provided to LLM to prevent hallucination when queries exceed available data boundaries.

\textbf{Task Decomposition and Planning.}
LLM-based task decomposition and planning \cite{Paranjape2023art, Li2024stride} leverages chain of thought (CoT) \cite{Wei2022COT} to ensure comprehensive instruction adherence and few-shot learning \cite{Brown2020fewshot} to guide the process. 
% For the data presenter it targets on enabling customized retrieval and aggregation, whereas for the insight generator, it involves selecting predetermined resolution paths based on domain following decomposition.
% Specifically, for the data presenter, by providing detailed descriptions of the available tool functionalities, the LLM is prompted to decompose the input query into a sequence of executable steps that can be accomplished using the specified tools, while for the insight generator, predetermined resolution paths encompassing data aggregation functions, analytical tools, and domain-aware knowledge are selected.
The data presenter enables customized data retrieval and aggregation through query decomposition into executable steps, by providing detailed API/function descriptions to LLM. The insight generator decomposes questions into domain-specific categories (such as performance, benchmarking, recommendation, etc) if necessary, and then selects predefined domain-aware resolution paths.

% An example of LLM-based task decomposition for data presenter can be found in Figure \ref{fig:decompose}. 
% To illustrate this step, use the question "\textit{How is my monthly sales doing year over year for the last month?}" as an example, which can not be answered via a single retrieval. After injecting the contextual information (current month is April), the original query is decomposed to two steps: (1) Retrieve monthly sales for the seller from March 2023 to March 2024 using data API, and (2) Calculate the year-over-year change in sales between March 2024 and March 2023 using calculation tool. For the insight generator, predetermined resolution paths encompassing data aggregation functions, analytical tools, and domain-aware knowledge are selected. The domain-aware knowledge contributes to constructing the prompt through instructions and multi-shot examples. For example, to answer the question "\textit{How is my business doing and how can I grow my business?}". The original query is decomposed to two steps: (1) Analyze business performance by utilizing performance report data, and (2) formulate recommendations by leveraging the recommendation data.

\textbf{API/function Selection with Payload Generation.}
The API and function selection is akin to LLM-based tool selection \cite{song2023restgpt, Qin2024toolllm}. 
% where the language model analyzes the given query and identifies the most suitable APIs or tools from the available repertoire to tackle the task effectively. 
The API/function name, description and column name are provided in the prompt to facilitate the process. This step essentially resembles the schema linking \cite{Katsogiannis2023text2sql}, i.e. table \& columns linking, where LLM creates the alignment between the entity references in the given query and the schema tables or columns. Subsequent to the API/function selection process, the next step involves payload/input generation with slot filling, where the required input parameters or arguments for the chosen tools or APIs are properly populated to facilitate their effective execution. In essence, the whole process is similar to code generation such as text-to-SQL,
% \cite{Gao2023texttosql, Li2023RESDSQL, Liu2023comprehensive, Pourreza2023DINSQL}, 
yet it is more robust in the sense that it is less prone to syntax errors and hallucinations (especially columns) associated with text-to-SQL.

\textbf{Domain-aware Routing for Insight Generator.}
% Within the insight generator agent, there is domain-aware branch routing process according to the relevant data sources based on the user query, achieved via few-shot learning based LLM classifier. In particular, the query could be routed to insights derived from domain-specific data. The routing process determines the resolution paths, which consist of the following: 
The insight generator also employs a few-shot learning based LLM classifier for domain-specific branch routing. The predetermined domain-aware resolution paths mainly contain (1) the associated analytical analysis
% that will be incorporated into the context, where analytical analysis 
that may include data aggregation, time series based seasonal \& trend analysis, as well as benchmark analysis among other techniques. (2) the domain-aware knowledge, prompt template and few-shot examples for the final insight generation.

% \begin{figure}
%   \centering
%   \includegraphics[width=10cm]{decompose.png}
%   \caption{Task decomposition by LLM}
%   \label{fig:decompose}
% \end{figure}
% \begin{figure*}[h]
%   \centering
%   \includegraphics[width=13.5cm]{insight_generator_gen_4.png}
%       \caption{Insight Generator Examples. Two distinct types of questions: business summary and benchmarking analysis are shown. The numbers provided are for illustrative purposes only and do not represent actual data.}
%   \label{fig:ig_gen}
% \end{figure*}
% \begin{figure}[h]
%   \centering
%   \includegraphics[width=8cm]{insight_generator_gen_4.png}
%       \caption{Insight Generator Examples. Two distinct types of questions: business summary and benchmarking analysis are shown. The numbers provided are for illustrative purposes only and do not represent actual data.}
%   \label{fig:ig_gen}
% \end{figure}

\subsubsection{Data Workflow Executor and Generation}
Data Workflow Executor component carries out data retrieval and data aggregation/transformation operations, adhering to the instructions outlined by the Data Workflow Planner. 
% For the insight generator, associated analytical tools such as time series based seasonal \& trend analysis and benchmark aggregation for benchmark data are also invoked as necessary. 
Additionally, data post processing tasks, such as reformatting the data, column renaming and semantic matching-based column filtering are also performed.
% , to reduce the context length, are also performed. The aforementioned method involves semantic matching-based column filtering using text embeddings.

The generation components mainly leverages in-context learning (ICL), with few-shot learning approach \cite{Brown2020fewshot} and CoT \cite{Wei2022COT}, to generate insights. The generation process for the data presenter is straightforward, with few-shot examples guiding the desired format of the response. Regarding the insight generator, domain experts provide specific domain knowledge and few-shot examples tailored to different data sources for LLM to follow.

\section{Experiments}\label{sec:exp}
% [TODO] Add the evaluation results from July launch

\subsection{Experimental Setup}
\subsubsection{Datasets.} To train OOD detection and agent routing models, we collected 301 commonly asked questions, with 178 in-domain and 123 out-of-domain respectively. The in-domain questions are further divided into 120 queries for the data presenter and 59 for the insight generator. To facilitate finetuning of the lightweight BERT model with a balanced dataset, the raw data presenter and insight generator questions are further augmented by LLM to be super-sampled to 300 questions each, introducing variations to expand the dataset. To evaluate IA end-to-end, a benchmarking dataset consisting of 100 carefully selected popular questions with ground truth are constructed.

\subsubsection{Setup.} The LLM employed in the experiments is based on "anthropic.claude-3-sonnet-20240229-v1:0" \cite{Anthropic2024claude} via Amazon Bedrock. Regarding the configuration of the OOD model, in our implementation the hyperparameter $\lambda$ is set as 4, and the dimension the hidden layer is 64. The base BERT model is "bge-small-en-v1.5" (33M parameters) \cite{bge_m3}.

\subsubsection{Metrics.}
The performance of OOD model and agent routing model are gauged via metrics such as precision, recall and accuracy. 
% The robustness of the data presenter is measured by precision and recall. Furthermore, 
Since there are multiple combinations of data retrieval/aggregation, we evaluate the retrieval performance together with end-to-end model performance, where the IA response is evaluated by human auditors on the benchmarking datasets. The response of IA are evaluated from the following three quantitative dimensions: 
% \begin{itemize}

% \item \textbf{Relevance}: Assess whether the key points and insights provided in the response directly address the question, which is defined as \textit{relevance = number of key words from the question that were addressed in the response / total number of key words in the question}

% \item \textbf{Correctness}: Verify the accuracy and reliability of the data and information presented in the response, which is similar to the notion of precision. \textit{correctness = number of correct insights in response / total number of insights in response}

% \item \textbf{Completeness}: Evaluate whether the response covers all the necessary data points or metrics that the user is inquiring about, which is akin to the concept of recall. \textit{completeness = number of required insights in response / total number of required insights}.
% \end{itemize}
\textbf{Relevance}: Assess whether the key points and insights provided in the response directly address the question, which is defined as \textit{relevance = \#\_key\_words from the question that were addressed in the response / total \#\_key\_words in the question.}

\textbf{Correctness}: Verify the accuracy and reliability of the data and information presented in the response (similar to the notion of precision). \textit{correctness = \#\_correct\_insights in response / total \#\_insights in response.}

\textbf{Completeness}: Evaluate whether the response covers all the necessary data points or metrics that the user is inquiring about (akin to the concept of recall). \textit{completeness = \#\_required\_insights in response / total \#\_required\_insights.}

The question-level accuracy is defined as \textbf{Question-level Accuracy}: the percentage of questions that have correctness, completeness, relevancy of more than 0.8.
% \begin{itemize}
% \item \textbf{Question-level Accuracy}: the percentage of questions that have correctness, completeness, relevancy of more than 0.8,
% \end{itemize}

\subsection{Experimental Results}\label{sec:eval}

\subsubsection{OOD Detection}

\begin{table}
  \caption{OOD Detection Performance}
  \label{ood-perf}
  % \centering
  \resizebox{\columnwidth}{!}{\begin{tabular}{llll}
    \toprule
    Model & Precision & Recall & Running Time (second) \\
    \midrule
    Auto-encoder & 0.969  & 0.721 & 0.009\\
    LLM-based few-shot & 0.616  & 0.971 & 1.665\\
    \bottomrule
  \end{tabular}}
\end{table}
% To test the accuracy of OOD module, we collect 301 sample queries with 172 positive (in-scope) and 129 negative (out-of-scope). 
% In our implementation, the hyperparameter $\lambda$ is set as 4, and the dimension the hidden layer is 64. 
As shown in table \ref{ood-perf}, AE-based OOD only takes less than 0.01s over each testing sample which significantly beats record compared to the LLM-based method. Meanwhile, it outperforms LLM significantly in terms of precision.
% in this high-precision classifier design since the purpose of OOD is to filter out questions that couldn't be addressed by IA quickly while ensuring coverage. 
As a further enhancement, the in-domain set (training set) could be enlarged to improve recall.

\subsubsection{Branch Routing} The results of branch routing is in table \ref{branch-rout}. Out of 178 in-domain samples, the model achieves classification accuracy of 0.83 with 0.3s latency for each routing case, compared with 0.60 accuracy in LLM-based classifier with $>$2s latency.

\begin{table}
  \caption{Branch Routing Performance}
  \label{branch-rout}
  % \centering
  \resizebox{\columnwidth}{!}{\begin{tabular}{lll}
    \toprule
    Model & Accuracy & Running Time (second) \\
    \midrule
    Finetuned BERT & 0.83  & 0.31 \\
    LLM-based few-shot & 0.60  & 2.14 \\
    \bottomrule
  \end{tabular}}
\end{table}

\subsubsection{Human Evaluation}
% Additionally, IA can also be evaluated from the following qualitative dimension:

% \textbf{Helpfulness}: Measure the clarity and usefulness of the insights provided in the response. This includes assessing whether the response uses 1) understandable language, 2) explains technical terms or metrics, 3) does not require follow-up clarification questions and 4) is useful for the seller.

For end-to-end IA response evaluation, the benchmarking dataset of 100 questions is sent to human auditors for evaluation providing the rubrics and ground truth. Out of 100 questions, 57 questions are in-scope, and the summary for the quantitative measures can be found in the table \ref{eval-human}.

\begin{table}
  \caption{Summary of evaluation metrics}
  \label{eval-human}
  \centering
  \resizebox{\columnwidth}{!}{\begin{tabular}{lllllll}
    \toprule
    % \multicolumn{1}{c}{}                   \\
    Metric & Avg & Std & Min & Max & Median & Samples \\
    \midrule
    Relevancy & 0.977  & 0.102 & 0.5 & 1  &1   &57 \\
    Correctness & 0.958 & 0.125 & 0.455 & 1   &1  &57 \\
    Completeness & 0.993 & 0.045 & 0.714 & 1  &1 &57\\
    \bottomrule
  \end{tabular}}
\end{table}

Meanwhile, the question-level accuracy is summarized in table \ref{acc}. The overall accuracy is high as 89.5\%. The end-to-end P90 latency is at 13.56s.

\begin{table}
  \caption{Question-level accuracy}
  \label{acc}
  \centering
  \begin{tabular}{lll}
    \toprule
    % \multicolumn{1}{c}{}                   \\
    Question-level Accuracy &Count &Percentage\\
    \midrule
    False & 6  & 10.5 \\
    True & 51 & 89.5 \\
    \bottomrule
  \end{tabular}
\end{table}

% To generate meaningful responses to better serve sellers, we plan to evaluate LLM responses from three angles: accurate factual information without hallucination (\textbf{reliability}), helpful responses that answer sellers' questions (\textbf{helpfulness}) without toxic, offensive or harmful language (\textbf{harmlessness}).

% While we try to develop a flexible and scalable evaluation framework, we rely heavily on human experts' feedback, especially regarding helpfulness and harmlessness, during the initial project phase. This domain-specific feedback from experts helps us further refine the responses to serve as golden requirements for our evaluation.

% Reliability evaluation is very close to the well-known Table-based Fact Verification NLP tasks, though existing benchmarking work are mostly based on public datasets (\citet{chen2020tabfact}, \citet{aly2021feverous}) and cannot be applied directly to our evaluation, as it is a very domain-specific task. To comprehensively evaluate generated insights and recommendations, we need to assess each metric numerical value within specific time periods, as well as evaluate descriptive elements such as trend directions and seasonal factors. This adds layer of complexity in evaluating both quantitative and qualitative aspects makes evaluation more intricate compared to document-based generations. 

\section{Conclusions} \label{sec:future}
%In this paper, we proposed a hierarchical master-worker multi-agent frameworkfor Insight Agents, which aims to help E-commerce sellers to grow their business via providing them personalized insight.  With more data sources becoming available and the need to cover a broader range of insight questions beyond business summaries, we aim to address the scalability challenges to enable seamless expansion and improved performance. Improving scalability for IA is important since 1) it enables the expansion to more use cases 2) it improves accuracy to our current solution and 3) it will support IA onboarding to other teams. The proposed solutions rely on RAG framework for dynamic retrieval and fine-tune LLM. For API selection, leverage a RAG framework to dynamically retrieve top K API’s for API selection step.
% : 1) Vector database to store API metadata with indexing to facilitate fast retrieval; 2) Semantic search for relevant API’s; 3) Reranking algorithm for final API’s. 
%For insight generation, 
% 1) Vector database to store knowledge base with indexing to facilitate fast retrieval, which includes instructions, background knowledge, multi-shot examples; 2) Semantic search for relevant instructions and examples; 3) Reranking algorithm for final instructions and examples; 4) 
%domain-aware prompt will be dynamically constructed.
% construction of insight generation prompt. Meanwhile, exploring text-to-SQL solutions can provide a more flexible approach for data aggregation. Another research area is automated evaluation techniques especially using LLM as the evaluator.

In this paper, we introduced Insight Agents (IA), a hierarchical multi-agent system leveraging LLMs built on a plan-and-execute paradigm to provide personalized, actionable insights for e-commerce sellers. Designed for comprehensive coverage, high accuracy, and low latency, our system significantly reduces cognitive load, achieving 89.5\% accuracy with sub-15-second latency. 
% By efficiently combining manager-worker agent framework, IA empowers sellers to optimize decision-making and drive growth. 
Looking ahead, we aim to scale IA to cover more use cases and integrate automated evaluation techniques for enhanced performance. Ultimately, IA sets the stage for future AI-driven decision supporting systems, transforming data interaction and driving impactful outcomes in e-commerce and beyond.

% \begin{figure}
%   \centering
%   \includegraphics[width=8cm]{data_presenter_gen_3.png}
%   \caption{Data Presenter Examples. (Artificial numbers for illustrative purposes only)}
%   \label{fig:dp_gen}
% \end{figure}

\section*{Appendix: sample outputs}
% Sample data presenter and insight generator answers can be found in Figure \ref{fig:ig_gen}.
\textit{Q: "What were the sales for my top 10 products last month?", A: "Your top 10 products by Sales for August 2024 (2024-08-01 ~ 2024-08-31) were: 1. Product1: \$30,000, 2. Product2: \$20,000, ..."}

\noindent
\textit{Q: "How is my business performing?"
, A: "In Feb 2024, your sales was \$10K (-\$6K or -50\% monthly YoY). Number of items sold were 200 (-50 or -20\% YoY). Average selling price came in at \$50 (-\$20 or -30\% YoY). Traffic was 30K views (+50\% YoY). Conversion rates were 0.50\% (-50 bps YoY). Overall business insights: ..."}

% \begin{figure}[h]
%   \centering
%   \includegraphics[width=7cm]{insight_generator_gen_4.png}
%       \caption{Insight Generator Examples. Two distinct types of questions: business summary and benchmarking analysis are shown. (Artificial numbers for illustrative purposes only)}
%   \label{fig:ig_gen}
% \end{figure}

% \begin{figure}[h]
%   \centering
%   \includegraphics[width=7cm]{data_insight_gen.png}
%       \caption{Sample output for data presenter and insight generator (artificial numbers for illustrative purposes only).}
%   \label{fig:ig_gen}
% \end{figure}

% \section*{PRESENTER BIOGRAPHY}
\section*{Presenter Biography}
% \textbf{Jincheng Bai} is an Applied Scientist at Amazon, focusing on Gen AI and Agentic LLM projects. He received his PhD in Statistics from Purdue University in US. 
% His research mainly focused on deep learning and natural language processing, and he has published at top-tier conferences including NeurIPS.

\textbf{Dr. Jincheng Bai} is a Machine Learning Scientist at Amazon, where he works on generative AI and agentic LLM systems. He earned his PhD in Statistics from Purdue University in the United States. His research focuses on deep learning and agentic AI, and he has published at top-tier conferences including NeurIPS and SIGIR.

\bibliographystyle{ACM-Reference-Format}
\balance
\bibliography{custom}

\end{document}